\documentclass[journal]{IEEEtran}

\usepackage{url}
\usepackage{array}
\usepackage{color}
\usepackage[dvipsnames]{xcolor}
\usepackage{amsmath}
\usepackage{amssymb}
\usepackage{graphicx}
\usepackage{afterpage}
\usepackage{algorithmic}
\usepackage{booktabs}
\usepackage[sort, numbers]{natbib}
\usepackage[caption=false,font=normalsize,labelfont=sf,textfont=sf]{subfig}

\newcommand{\etal}{\MakeLowercase{\textit{et al.}}}

\begin{document}

\title{Uncertainty-Aware Deep Ensembles for Reliable and Explainable Predictions of Clinical Time Series}

\author{Kristoffer~Wickstrøm,
        Karl~Øyvind~Mikalsen,
        Michael~Kampffmeyer,
        Arthur~Revhaug,
        and Robert Jenssen% <-this % stops a space
\thanks{K. Wickstrøm, KØ. Mikalsen, M. Kampffmeyer, and R. Jenssen are with the UiT Machine Learning Group at the Dept. of Physics and Technology, UiT the Arctic University of Norway, Norway,
Tromsø, NO-9037, e-mail: kwi030@uit.no.}%
\thanks{A. Revhaug is with the Dept. of Clinical Medicine, UiT the Arctic University of Norway, Tromsø, Norway}%
\thanks{KØ. Mikalsen is also with the Dept. of Gastrointestinal Surgery, University Hospital of North Norway (UNN), Tromsø, Norway}}

\markboth{}%
{Shell \MakeLowercase{\textit{et al.}}: Bare Demo of IEEEtran.cls for IEEE Journals}

% The only time the second header will appear is for the odd numbered pages
% after the title page when using the twoside option.
% 
% *** Note that you probably will NOT want to include the author's ***
% *** name in the headers of peer review papers.                   ***
% You can use \ifCLASSOPTIONpeerreview for conditional compilation here if
% you desire.

% If you want to put a publisher's ID mark on the page you can do it like
% this:
%\IEEEpubid{0000--0000/00\$00.00~\copyright~2015 IEEE}
% Remember, if you use this you must call \IEEEpubidadjcol in the second
% column for its text to clear the IEEEpubid mark.

\maketitle

\begin{abstract}
Deep learning-based support systems have demonstrated encouraging results in numerous clinical applications involving the processing of time series data. While such systems often are very accurate, they have no inherent mechanism for explaining what influenced the predictions, which is critical for clinical tasks. However, existing explainability techniques lack an important component for trustworthy and reliable decision support, namely a notion of uncertainty. In this paper, we address this lack of uncertainty by proposing a deep ensemble approach where a collection of DNNs are trained independently. A measure of uncertainty in the relevance scores is computed by taking the standard deviation across the relevance scores produced by each model in the ensemble, which in turn is used to make the explanations more reliable. The class activation mapping method is used to assign a relevance score for each time step in the time series. Results demonstrate that the proposed ensemble is more accurate in locating relevant time steps and is more consistent across random initializations, thus making the model more trustworthy. The proposed methodology paves the way for constructing trustworthy and dependable support systems for processing clinical time series for healthcare related tasks.
\end{abstract}

\begin{IEEEkeywords}
deep learning, ensembles, interpretability, uncertainty, time series
\end{IEEEkeywords}

% For peer review papers, you can put extra information on the cover
% page as needed:
% \ifCLASSOPTIONpeerreview
% \begin{center} \bfseries EDICS Category: 3-BBND \end{center}
% \fi
%
% For peerreview papers, this IEEEtran command inserts a page break and
% creates the second title. It will be ignored for other modes.
\IEEEpeerreviewmaketitle

\section{Introduction}

\IEEEPARstart{C}{linical} data stored in electronic health records (EHRs) contain valuable information that can be used for e.g. diagnosis support \cite{8918246}. The type of data stored in EHRs can vary between a number of different modalities, for instance free text (e.g. nursing notes) or clinical time series (e.g. blood or temperature measurements). Recent advances in machine learning have shown how such information can be extracted from EHRs and used to construct data-driven algorithms, which can serve as support systems that aid medical practitioners in decision making \cite{HrayrMIMIC}. Particularly, systems based on deep neural networks (DNNs) have shown promising results on a number of tasks such as mortality prediction \cite{HrayrMIMIC}, detection of infections \cite{andreas}, and patient treatment trajectory prediction \cite{8941234}.

While DNNs often provide accurate predictions, they have no inherent mechanism for explaining what influenced the predictions. This has been noted on numerous occasions, and has resulted in DNNs often being referred to as black-boxes \cite{Alain2017UnderstandingIL}. A recent study found that it is crucial to; 1) know the subset of features deriving the model outcome and 2) provide a measure of uncertainty for predictions for creating trustworthy machine learning-based support systems \cite{Tonekaboni2019WhatCW}. These mechanisms are not built into DNNs, but recent advances in explainable artificial intelligence (XAI) have made great leaps in developing methods that provide interpretations for the prediction of a model \cite{Samek2017ExplainableAI}.

\begin{figure*}[ht]
    \centering
    \includegraphics[width=0.975\textwidth]{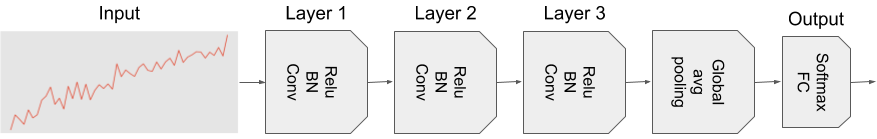}
    \caption{Figure illustrates the architecture of the FCN proposed by Wang \etal\ (2017) \cite{7966039}. Layer 1, 2, and 3 are convolutional blocks that consists of a convolutional operation, followed by batch normalization and a ReLU activation function. Block 1, 2, and 3 contains 128, 256, and 128 filters of size 7, 5, and 3, respectively. The three convolutional blocks are followed by a global average pooling layer that averages over the time dimension of the out output of the last convolutional block. Lastly, the output layer of the model is a fully connected layer followed by a softmax activation function. At each convolutional block, the input is zero-padded such that the length of time series does not diminish. This makes FCNs particularly suited for utilizing the CAM technique for explainable CNNs, as a relevance score can be computed for each time step in the time series.}
    \label{fig:fcn}
\end{figure*}

One promising method for XAI is based on the so-called class activation mapping (CAM) method \cite{7780688}. CAM was originally developed for image data, but has also been shown to be applicable for temporal data such as clinical time series \cite{7966039, XAIIJCAI}. By utilizing DNNs with a particular processing structure, CAM can assign a score to each time step in an input that indicates how important that time step is for a given prediction. This score will be referred to as a relevance score. Explainable methods that provide a notion of uncertainty are lacking, but needed to provide trustworthy and dependable support systems. For instance, if a clinical measurement is identified as being highly relevant for a prediction, how certain is this relevance score? Or if several DNNs are trained from different random initialization, will the same measurements be highlighted as relevant across the different models? Such questions cannot be answered within the standard framework of DNNs. If explanations are accepted without taking uncertainty into account it might results in an unjustified belief in the explanations.

In this work we propose a deep ensemble approach to model uncertainty in explainability for DNN-based predictions of clinical time series. A collection of DNNs are trained independently, each producing a prediction and a relevance score for each time step. The uncertainty in the relevance scores is computed by taking the standard deviation across the relevance scores produced by each model in the ensemble. Intuitively, time steps that all models indicate as relevant will be highlighted as certainly relevant. Time steps that only one or some models highlight as relevant will be highlighted as relevant, but with a high degree of uncertainty. To the best of our knowledge, such a deep ensemble approach for uncertainty in explainable DNNs for clinical tasks has not been previously explored. The proposed approach is validated on synthetic data and on two clinically relevant tasks; myocardial infarction detection in echocardiograms (ECGs) and surgical site infection (SSI) in blood measurements of C-reactive protein (CRP). Experiments show how the deep ensemble is more accurate at locating relevant time steps and more consistent. Consistent means that the model highlights similar time steps as relevant when retrained from different initializations. Further, the value of the uncertainty measures obtained by the proposed methodology is demonstrated through several qualitative experiments. Although in this work the method is illustrated with the CAM approach, it is applicable for any explainability technique for DNNs. The proposed approach paves they way for increasing trustworthiness of DNNs and can be an important component in constructing dependable and transparent decision support systems.

\section{Related Work}

Recently there has been a great increase in methods for creating explainable DNNs. This section will describe the recent works that are most closely related to this paper. For a more comprehensive review the reader is referred to a summary of XAI by Samek \etal\ (2017) \cite{Samek2017ExplainableAI} and an overview of XAI in healthcare by Tonekaboni \etal\ (2019) \cite{Tonekaboni2019WhatCW}.

Many approaches have been proposed to create interpretable decision support systems based on DNNs. Zhang \etal\ (2018) proposed a model based on recurrent neural networks (RNNs) that could learn an interpretable deep representation that was personalized for each patients \cite{Zhang2018Patient2VecAP}. This was achieved through the attention mechanism \cite{attention}, which was used to indicate the relative importance of different features to the personalized embedding of a patient. Assaf and Schumann (2019) proposed a gradient-based interpretablity approach for convolutional neural networks (CNNs) that handles multivariate time series through a two-stage approach \cite{XAIIJCAI}. They demonstrate how their approach can be used to explain which features during a time interval are important for a given prediction. Tonekaboni \etal\ (2020) introduced a method that automatically assigns an importance values to each features at each time step by simulating counterfactual trajectories given previous observations \cite{tonekaboni2020explaining}. However, apart from Wickstr\o m \etal\ (2020) \cite{WICKSTROM2020101619} who proposed an approach for providing uncertainty measures for input feature importance in computer vision tasks, the issue of uncertainty in input feature importance have, to the best of our knowledge, not been explored in the context of EHRs.

\section{Fully Convolutional Networks}
\label{sec:FCN}

Several recent works have shown that CNNs can achieve state-of-the-art performance on time series classification tasks, and are usually easier to train than RNNs \cite{7966039, Fawaz2019InceptionTimeFA}. In this paper we use a network similar to the fully convolutional network (FCN) proposed by Wang \etal\ (2017) \cite{7966039}, which has demonstrated good performance on time series classification benchmarks \cite{7966039}. The FCN consists of three convolutional blocks, each consisting of a convolution operation, batch normalization \cite{batchnorm} and a rectified linear unit (ReLU), a global average pooling operation and a fully connected layer followed by a softmax activation function for the output layer. The convolutional blocks and the pooling layer can be considered the encoder part of the model, while the fully connected layer combined with the softmax function constitute the classifier of the model. An illustration of the model is shown in Figure \ref{fig:fcn}. The first convolutional block consists of 128 filters with size 7, the second of 256 filters with size 5, and the third convolutional block has 128 filters with size 3. An important component of the FCN is that the input is zero-padded such that the length of the time series does not change through the three convolutional blocks. This is vital for the interpretablity technique that is discussed in Section \ref{sec:CAM}. The global average pooling operation, which takes the average over the entire time dimension, summarizes the content of the filtered time series and produces a single value for each filter in the last convolutional block.

\section{Class Activation Mapping}
\label{sec:CAM}
\begin{figure*}[ht]
    \centering
    \includegraphics[width=0.975\textwidth]{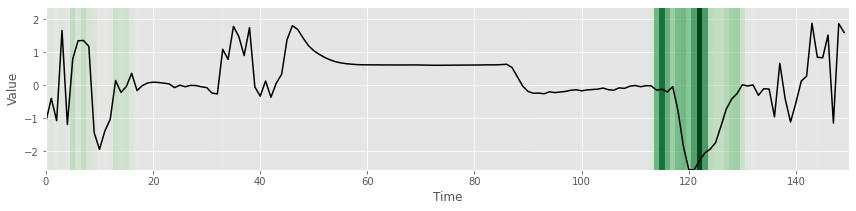}
    \caption{Illustration of CAM technique for CNN interpretablity. Green indicates positive relevance to the prediction. The figure displays an example of a time series that is characterized by a downward dip in the initial of final periods of the time series. This example is correctly classified by the FCN, and the CAM highlights the time steps associated with the downward dip towards the end of the time series as the most relevant features for the prediction.}
    \label{fig:umdEx}
\end{figure*}

CAM is an interpretability method designed for CNNs that highlights class-specific relevance scores in the input data \cite{7780688}. CAM have recently shown encouraging results for tasks involving non-clinical and clinical time series \cite{7966039, XAIIJCAI}. Let $w_{c,k}$ denote the weight connecting the $k^{th}$ filter in the last convolutional block with the neuron corresponding to class $c$ in the output layer, and $z_{k,t}$ denote the activation at time step $t$ produced by the $k^{th}$ filter in the last convolutional block of the FCN. Then, the input to the final softmax function in the output layer can be expressed as:

\begin{equation}\label{eq:CAM}
    g_c = \sum\limits_{k=1}^{K} w_{c,k}\sum\limits_{t=1}^T z_{k, t} = \sum\limits_{t=1}^T r_{c,t},
\end{equation}
where $K$ is the number of filters in the last convolutional block,  $T$ is the length of the time series, and $r_{c, t}$ denotes the relevance score of time step $t$ for class $c$:
\begin{equation}\label{eq:rel}
    r_{c,t} = \sum\limits_{k=1}^{K} w_{c,k} z_{k, t}.
\end{equation}

Equation \ref{eq:CAM} and \ref{eq:rel} show that the relevance score at a given time step can be directly related to the input of the softmax function in the output layer, i.e. what produces the prediction of the model. Equation \ref{eq:rel} also illustrates why the zero-padding in the FCN makes the architecture particularly suited for CAM. Since the output of the last convolutional block has the same length as the input, a relevance score for each time step can be computed. The CAM method can be understood as a weighted linear sum of the presence of different patterns in the input data, which can be used to identify the input regions most relevant to the particular category \cite{7780688}. While it is possible to also use CAM with CNNs that reduce the length of the time series during processing through different upsampling procedures, it does complicate the procedure significantly. Furthermore, many interpretability techniques only consider the features that have a positive relevance for a given prediction, for instance the guided backpropagation method \cite{springenberg2014striving}. This is to provide clearer and unambiguous explanation for a prediction. While negative relevance scores can in some cases provide additional information, they can also be difficult to interpret. In binary classification problems, such as those considered in this work, the positive class typically makes up a homogeneous group with some shared defining characteristic (e.g. an elevate blood measurement). In contrast, the negative class (e.g. control patients) can be very different and typically makes up a highly heterogeneous group that can be difficult to interpret. Removing all negative relevance scores can be achieved by modifying Equation \ref{eq:rel} as follows:

\begin{equation}\label{eq:relPluss}
    r_{c,t} = \text{max}\Big(0, \sum\limits_{k=1}^{K} w_{c,k} z_{k, t}\Big).
\end{equation}

To end this section, an example illustrating the CAM approach is presented.
The CAM interpretability technique is illustrated by training a FCN on the UMD dataset \cite{UAEUTS}. This dataset has three classes, one characterized by a downward dip in the initial or final period of the time series, one characterized by an upward dip in the initial or final period of the time series, and one characterized by no dip. Figure \ref{fig:umdEx} shows an example that belongs to the first class, because of the significant dip towards the end of the time series, and is correctly classified by the FCN. Figure \ref{fig:umdEx} displays the relevance scores produced for the first class by the CAM, where green indicates that a time step is relevant for classifying the sample to the first class. Figure \ref{fig:umdEx} clearly indicates that the FCN is focusing on the downward dip towards the end of the time series, which is the defining characteristic of this class.

\begin{figure*}[ht]
    \centering
    \includegraphics[width=\textwidth]{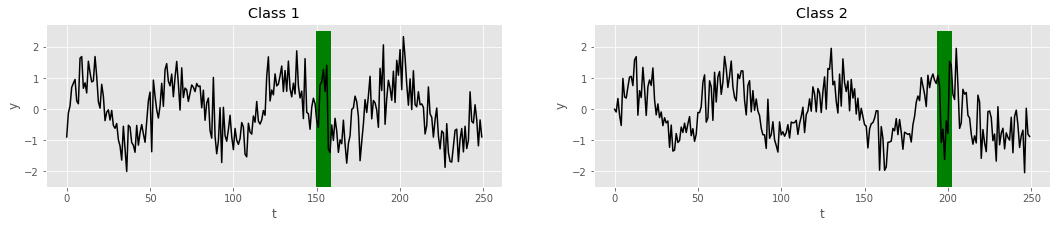}
    \caption{Illustration of the synthetic data constructed for quantitative analysis of the proposed ensemble method. The leftmost plot shows an example from class 1, where an upwards spike is the defining characteristic. The rightmost plot shows an example from class 2, where a downward spike is the defining characteristic. The characteristic spikes are highlighted in the plots.}
    \label{fig:synthEx}
\end{figure*}

\section{Ensembles for Uncertainty Estimation in Explainability}

In this work, we propose an ensemble of FCNs for classification of EHRs. An ensemble is comprised of a set of separately trained classifiers which are combined to provide predictions when presented with new data. Ensemble approaches have been widely used in machine learning \cite{ens1, ens2} as they offer a number of advantageous properties. Ensembles are typically more accurate, consistent and have lower variance than their single model counterparts \cite{ens1, ens2}. Furthermore, they enable estimation of uncertainty in predictions and are simple and applicable for most tasks \cite{ens1, ens2}. The limitation of ensembles approaches are usually computational, but in the context of EHRs classification data is often limited and smaller models can usually be utilized \cite{missdata}. Previous work on deep ensembles have demonstrated how they can increase accuracy and provide reliable uncertainty estimates \cite{ensemble}. Furthermore, as the amount of data can be limited for medical tasks \cite{missdata}, the training procedure can be unstable. However, an ensemble of classifiers are known to tackle such issues well \cite{Leobagging}. The uncertainty measures are obtained by computing the standard deviation across the relevance scores for each model in the ensemble.

To calculate the ensemble mean and standard deviation of the relevance scores, let each CNN be parametrized by a set of parameters $\left\{\theta_1, \cdots ,\theta_M\right\}$, where $M$ is the number of models in the ensemble. Then, when considering the positive relevance scores from Equation \ref{eq:relPluss}, the mean relevance across the ensemble is defined as:

\begin{equation}\label{eq:muR}
    \boldsymbol{\mu}_{\mathbf{r}} = \frac{1}{M}\sum\limits_{m=1}^{M} \mathbf{r}_{\theta_m},
\end{equation}

and the standard deviation across the ensemble is defined as:

\begin{equation}\label{eq:stdR}
    \boldsymbol{\sigma}_{\mathbf{r}} = \sqrt{\frac{1}{M-1}\sum\limits_{m=1}^{M} (\mathbf{r}_{\theta_m}-\boldsymbol{\mu}_{\mathbf{r}})^2}.
\end{equation}

In Equation \ref{eq:muR} and \ref{eq:stdR}, $\mathbf{r}_{\theta_m}$ refers to the relevance scores for a given sample $\mathbf{x}$ provided by the model parametrized by the parameter set $\theta_m$. Note that relevance scores should be scaled to the same range for them to be comparable. In all experiments we scale the relevance scores between 0 and 1 using the following equation:

\begin{equation}
    \tilde{\mathbf{r}}_c = \frac{\mathbf{r}_c-\text{min}(\mathbf{r}_c)}{\text{max}(\mathbf{r}_c)-\text{min}(\mathbf{r}_c)}.
\end{equation}

\subsection{Uncertainty Filtered Relevance Scores}\label{sec:UFRS}

For challenging datasets, several models in the ensemble might disagree on which time steps are most relevant for a prediction. In such cases, it might be desirable to only consider the time steps that most models in the ensemble agrees on, i.e. to filter out the uncertain relevance scores. With that in mind, a modification of the relevance scores is proposed, which will be referred to as uncertainty filtered relevance scores. For a given sample, the uncertainty filtered relevance scores are calculated by considering the standard deviation for each time step across all models in the ensemble. If the standard deviation is below some threshold the relevance scores are kept as it is. If the standard deviation is above some threshold the relevance scores are set to zero. This can be formulated as:

\begin{equation}\label{eq:muRstd}
    \tilde{\mu}_{r_t} = \begin{cases}
\mu_{r_t}&, \text{ if }  \sigma_{r_t} < \epsilon\\
0&, \text{ else}
\end{cases}.
\end{equation}

The threshold, $\epsilon$, can be picked to fit the particular data in question. If only time steps with high certainty should be considered, a low threshold can be chosen. Or if uncertainty is not a concern, a high threshold can be selected. For this work, a simple but intuitive heuristic is proposed for setting the value of $\epsilon$. Let the threshold be set to the mean of the standard deviation across all time steps of a given samples, that is:

\begin{equation}\label{eq:epsilon}
    \epsilon = \frac{1}{T}\sum\limits_{t=1}^T \sigma_{r_t}.
\end{equation}

This approach will ensure that the most uncertain relevance scores will be filtered out and will adapt the threshold to each specific sample.

\section{Synthetic Data for Quantitative Assessment of Explainability}\label{sec:synthData}

A challenging aspect of XAI is that it is inherently qualitative, which makes quantitative assessment difficult. This is because for real datasets it is rarely known exactly what time steps are important. Therefore, to evaluate the proposed methodology, a synthetic dataset is constructed in such a way that the relevant time steps are known in advance. A time series classification task with two classes is constructed using the Python TimeSynth package\footnote{\url{https://github.com/TimeSynth/TimeSynth}}, following the example of Tonekaboni \etal\ (2020) \cite{tonekaboni2020explaining}. The data is constructed to resemble the periodic nature of ECG measurements in the presence of noise. Class number one is characterised by an upward spike, and class number two is characterised by a downward spike. The spike spans five time step, which are labeled as relevant time steps for the sample. Additionally, the two time steps preceding and succeeding the spike are also labeled as relevant time steps. These time steps are also chosen to be relevant as the change from no-spike to spike and vice versa is also important for the model to pick up. In total, there are 9 relevant points in each time series. Each time series consists of 250 time steps sampled from a univariate sinusoidal wave with a frequency of 0.2. Gaussian noise with zero mean and a variance of 0.5 is added at each time step. An example of the data and each class can be seen in Figure \ref{fig:synthEx}.

\paragraph{Calculating relevance accuracy} To calculate how accurate a model is a locating relevant time steps, we compare the $k$ most relevant time steps for a prediction with the known most relevant times steps. For a given sample $i$, let $Y_i=\{y_1^{(i)}, \cdots, y_{N_r}^{(i)}\}$ denote the set of relevant points, where $N_r$ is the number of known relevant time steps in the time series, and $R_i(k)$ denote the $k$ most relevant time steps for the prediction of the model. The order of the relevance scores is not taken into account here, as all the relevant points are considered equally important in this case. Perfect relevance accuracy is achieved when all elements of $Y_i$ are contained in $R_i(k)$, preferably with as few $k$s as possible. For a given sample $i$, Relevance accuracy can be calculated by dividing the cardinality of the intersection of $R_i(k)$ and $Y_i$ by the cardinality of $Y_i$, which can be expressed as:

\begin{equation}\label{eq:accuracy}
    \text{Relevance accuracy} = \frac{\lvert R_i(k) \cap Y_i \rvert}{\lvert Y_i \rvert},
\end{equation}

where $\cap$ is the intersection of two sets and $\vert \cdot \rvert$ is the cardinality of a set. Note that the relevant points are ranked from least to most relevant before they are compared to $Y_i$. Therefore, simply highlighting all time steps as relevant will not result in a high relevance accuracy score, the model needs to highlight some time steps as being more important than the others.

\paragraph{Calculating relevance consistency} For a model to be trustworthy, it should indicate mostly the same time steps as relevant for its prediction when trained from a different initialization. Models that highlight the same time steps as relevant when trained from different initialization will be referred to as consistent. Relevance consistency can be calculated in a similar way as relevance accuracy was computed. For a given sample, let $R_i^{(m)}(k)$ and $R_i^{(n)}(k)$ denote the $k$ most relevant time steps for the prediction of two models trained from different initialization. Relevance consistency is computed by counting the number of shared elements of the two sets. As with relevance accuracy, the order is not taken into account. This computation can be formulated as, for a given sample, to compute the cardinality of the intersection of the set of the $k$ most relevant time steps for the prediction of two models for then to divide by $k$. For a given sample $i$, computing the relevance consistency across $M$ models can then be achieved by:

\begin{equation}\label{eq:consistency}
    \text{Relevance consistency} = \frac{1}{M}\sum \limits_{m,n=1}^{M}\frac{\lvert R_i^{(m)}(k) \cap R_{i}^{(n)}(k) \rvert}{k}.
\end{equation}

Similarly as for Equation \ref{eq:accuracy}, the ranking of the relevance scores is important. To achieve high relevance consistency, the model must highlight the same time steps as more relevant than other time steps, even when trained from a different random initialization.

\section{Experiments and Discussion}

Several experiments are conducted that demonstrate the benefits of the proposed methodology for creating explainable support systems based on DNNs. First, the proposed approaches are validated on synthetic data. Next, the relevance consistency of the ensemble approach is validated on both synthetic and clinical data. Further, the proposed approaches are used to determine what inputs are important for classifying ECGs as a normal heartbeat or a myocardial infarction (heart attack). A similar experiment is also conducted for identifying patients with surgical site infection in blood measurements of CRP. Table \ref{tab:DataDesc} provides an overview of the different properties of all datasets used to evaluate the proposed methodology. The table shows, for each dataset, the number of samples, the class distribution, the length of the time series, what each time step represents and what type of measurement that is considered. The FCN described in Section \ref{sec:FCN} is used for all tasks, and is trained using a cross-entropy loss and the Adam optimizer \cite{ADAM}. For all experiments in this section, all ensembles are composed of 10 FCNs. To evaluate the performance of the classifiers we compute four metrics on the test data of each dataset; precision, recall, negative predictive value (NPV), and specificity. These metrics are chosen to reflect typical challenges when evaluating performance in classification of clinical time series, such as unbalanced data and false positives. The code used in this manuscript is available at \url{https://github.com/Wickstrom/TimeSeriesXAI}.

\begin{table}[ht]
\centering
\caption{Description of the three datasets used in experiments, including the number of training and test samples (N$_\text{tr}$, N$_\text{te}$), the length of the time series (T), what each time step represents, and the type of data (Data). The table also displays the number of samples for each class (C=0 and C=1) in both the training and test data.}
\begin{tabular}{l|c|c|c}
 &
\multicolumn{1}{c|}{\textbf{Synthetic}} &
\multicolumn{1}{c|}{\textbf{ECG200}} &
\multicolumn{1}{c}{\textbf{SSI}} \\
\toprule
N$_\text{tr}$ (C=0) & 250 & 69 & 520 \\
\hline
N$_\text{tr}$ (C=1) & 250 & 31 & 185 \\
\hline
N$_\text{te}$ (C=0) & 250 & 64 & 130 \\
\hline
N$_\text{te}$ (C=1) & 250 & 36 & 48 \\
\hline
T & 250 & 96 & 20 \\
\hline
TS & $\cdot$ & microseconds & days \\
\hline
Data & $\cdot$ & heartbeat & C-reactive protein \\
\bottomrule
\end{tabular}
\label{tab:DataDesc}
\end{table}

\subsection{Validation on Synthetic Data}

Following the procedure described in Section \ref{sec:synthData}, a training set of 500 samples is generated, along with a separate test set of 500 samples. The performance of a single FCN and an ensemble of FCNs on the synthetic test data, each trained for 150 epochs, is displayed in Table \ref{tab:metrics}. A Monte Carlo permutation test with 10000 permutations is conducted to test for significance. Results indicate that the ensemble has higher precision and produce less false positives. 

\begin{table*}[ht]
\centering
\caption{Evaluation of classification performance of single and ensemble model on the test data of different dataset. Bold number indicate statistical significance at a significance level of 0.01.}
\begin{tabular}{l|c|c|c|c|c|c|c|c}
\multicolumn{1}{c|}{\textbf{Dataset}} &
\multicolumn{2}{c|}{\textbf{Precision}} &
\multicolumn{2}{c|}{\textbf{Recall}} &
\multicolumn{2}{c|}{\textbf{NPV}} &
\multicolumn{2}{c}{\textbf{Specificity}} \\
\toprule
& Single & Ensemble & Single & Ensemble & Single & Ensemble & Single & Ensemble \\
\hline
Synthetic & .983$\pm$.011 & \textbf{.991$\pm$.001} & .968$\pm$.015 & .962$\pm$.002 & .968$\pm$.014 & .963$\pm$.002 & .983$\pm$.028 & \textbf{.992$\pm$.001} \\
\hline
ECG200 & .819$\pm$.018 & .814$\pm$.023 & .741$\pm$.044 & \textbf{.755$\pm$.035} & .862$\pm$.020 & .867$\pm$.016 & .908$\pm$.010 & .903$\pm$.014 \\
\hline
SSI & .947$\pm$.069 & \textbf{.978$\pm$.015} & .922$\pm$.037 & .936$\pm$.041 & .973$\pm$.013 & .976$\pm$.013 & .981$\pm$.029 & .993$\pm$.006 \\
\bottomrule
\end{tabular}
\label{tab:metrics}
\end{table*}

\paragraph{Relevant time steps accuracy} The relevance accuracy of the single and ensemble model for different values of $k$ are presented in Table \ref{tab:SyntheticResults}, where the results are averaged over 10 independent training runs. As described in Section \ref{sec:synthData}, the data is constructed in such a way that there are 9 time steps that are considered relevant, inspired by Tonekaboni \etal\ (2020) \cite{tonekaboni2020explaining}. Therefore, we start by evaluating the relevance accuracy at $k=9$ and above. Table \ref{tab:SyntheticResults} shows that the FCN is capable of identifying the relevant samples in time series with high relevance accuracy. Furthermore, the results show that the deep ensemble is more accurate at identifying relevant samples compared to single models, and also has much less variability in their prediction. A Monte Carlo permutation test with 10000 permutations is conducted to test for significance, and the difference between the single and the ensemble model is significant for most $k$s at a significance level of 0.01.

\paragraph{Most highlighted relevant time steps} A priori, it is not obvious which time steps the model will highlight most frequently as being relevant for predictions. However, it is desirable that the model use the known relevant time steps for its prediction. To evaluate which time steps are most frequently used to make a prediction, we define a relevance ratio (RR). For a given sample $i$ and a known relevant time step $y_j^{(i)}$, the RR is expressed as:

\begin{equation}\label{eq:ratio}
   RR_j =  \frac{1}{N_{te}}\sum\limits_{i=1}^{N_{te}} I(y_j^{(i)}, R_i(k)),
\end{equation}

where $I$ is an indicator function that evaluates to 1 if the known relevant sample $y_j^{(i)}$ is among the k most relevant points $R_i(k)$ and $N_{te}$ is the number of test samples. If $RR_j=1$, then the known relevant time step $y_j^{(i)}$ is included in $R_i(k)$ for all samples. It is expected that a random time step will be included $N_{te}(k/T)$ number of times among the most relevant time steps, which can be considered a lower bound for how many times a time step should be included.

\begin{figure}[ht]
    \centering
    \includegraphics[width=0.475\textwidth]{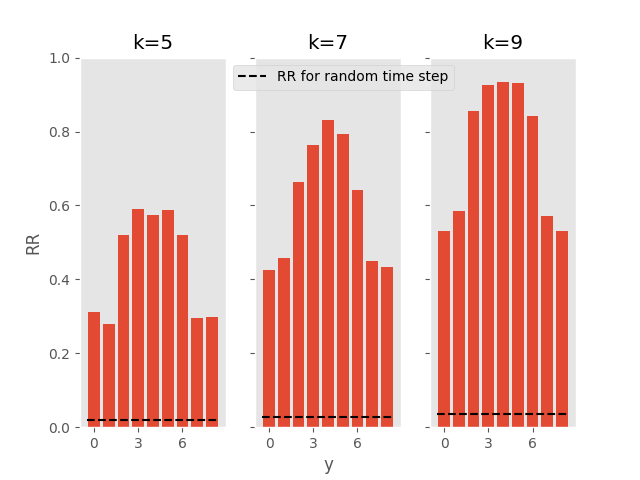}
    \caption{Figure shows the number of times a known relevant time step is included in the top k most relevant time steps across all test samples in the synthetic dataset. Results indicate that the time steps at the middle of the spike is the most frequently highlighted as relevant for a prediction, and that the known relevant time steps are more frequently highlighted as being important compared to non-relevant time steps.}
    \label{fig:barplotRel}
\end{figure}

Figure \ref{fig:barplotRel} displays which time steps are most frequently highlighted as relevant for different values of $k$. In Figure \ref{fig:barplotRel}, the two initial and two final bars correspond to the two initial and two final relevant time steps, while the five bars in the middle correspond to the spike in the synthetic data. Results indicate that the five time steps corresponding to the spike are most frequently highlighted as being relevant for the prediction. The figure also shows that the ensemble model uses the known relevant time steps much more frequently than other non-relevant time steps.

\begin{table}[ht]
\centering
\caption{Relevance accuracy of FCN for relevance scores on synthetic data. Scores are averaged over 10 runs for both the single and ensemble model. Bold number indicate statistical significance at a significance level of 0.01.}
\begin{tabular}{c|l|l}
\multicolumn{1}{c|}{\textbf{Top k}} &
\multicolumn{1}{c|}{\textbf{Single}} &
\multicolumn{1}{c}{\textbf{Ensemble}} \\
\toprule
k=9 & .703 $\pm$ 0.014 & \textbf{.736 $\pm$ .014} \\
\hline
k=10 & .765 $\pm$ 0.013 & \textbf{.800 $\pm$ .011} \\
\hline
k=11 & .821 $\pm$ 0.012 & \textbf{.855 $\pm$ .011} \\
\hline
k=12 & .868 $\pm$ 0.010 & \textbf{.899 $\pm$ .007} \\
\hline
k=15 & .935 $\pm$ 0.007 & \textbf{.950 $\pm$ .001} \\
\bottomrule
\end{tabular}
\label{tab:SyntheticResults}
\end{table}

\subsection{Relevance consistency in Relevance Scores}

\begin{figure*}[ht]
\centering
  \subfloat[\label{fig:singleConsistency}]{%
       \includegraphics[width=0.95\linewidth]{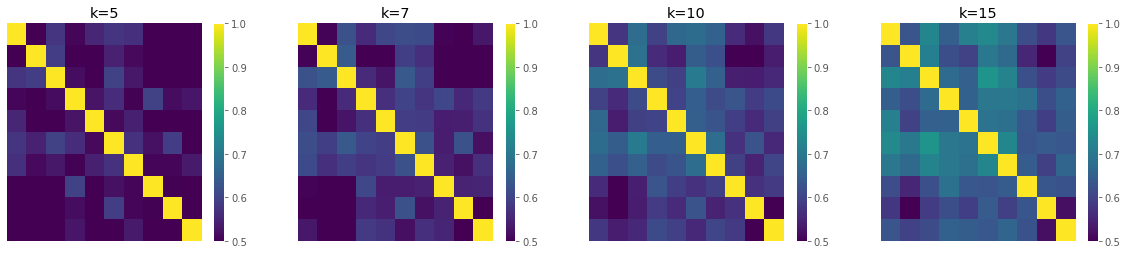}}
    \hfill
  \subfloat[\label{fig:ensembleConsistency}]{%
       \includegraphics[width=0.95\linewidth]{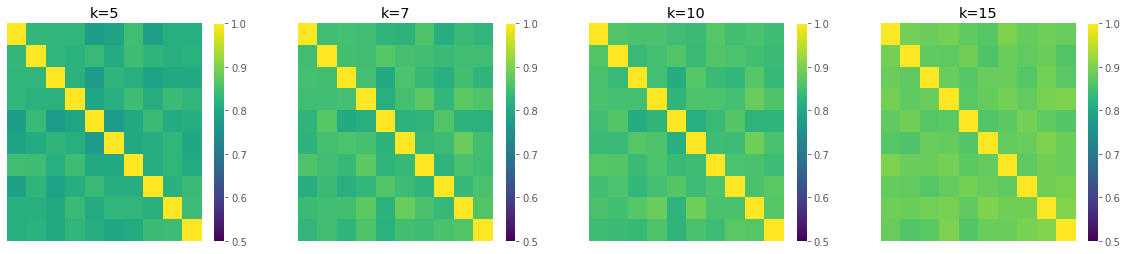}}
    \hfill
\caption{Relevance consistency across 10 single and ensemble models for different number of $k$ relevant time steps on the ECG200 test data. Top row shows relevance consistency of single models and bottom row shows relevance consistency in ensemble models. Figure displays how ensemble models regularly highlight the same time steps as relevant for their prediction, while the single models have much more variability in what time steps are being indicated as relevant for their prediction.}
\label{fig:consistency}
\end{figure*}

The relevance consistency of single and ensemble models is evaluated on both the synthetic data described in the previous section and on the ECG200 dataset \cite{olszewski2001generalized, UAEUTS}. The ECG200 dataset consists of ECGs that traces the electrical activity recorded during a single heartbeat. It is obtained from the UCR time series classification archive \cite{UAEUTS} and has a predefined training and test split. The task is to discriminate between normal heartbeats and those associated with myocardial infarction, also known as a heart attack. Table \ref{tab:Consistency} shows the relevance consistency of 10 single and 10 ensemble models on both datasets. Results demonstrate that the ensemble approach is far more consistent than the single models, and has much lower variability in its scores. This is particularly prominent for the more challenging ECG200 data, where the single models have difficulties with agreeing on the what time steps are relevant compared to the ensemble approach. A Monte Carlo permutation test with 10000 permutations is conducted to test for significance in both datasets, and the difference between the single and the ensemble model is significant for all $k$s at a significance level of 0.01. Furthermore, Figure \ref{fig:consistency} shows the relevance consistency between the 10 single models and 10 ensemble models on the ECG200 dataset. The figure corroborate the quantitative results in Table \ref{tab:Consistency} that show how the ensemble approach is more consistent than single models. 

\begin{table}[ht]
\centering
\caption{Relevance consistency of relevance scores averaged over 10 single and 10 ensemble models on synthetic data and ECG time series. Bold number indicate statistical significance at a significance level of 0.01.}
\begin{tabular}{c|l|l|l|l}
\multicolumn{1}{c|}{\textbf{Top k}} &
\multicolumn{2}{c|}{\textbf{Synthetic}} &
\multicolumn{2}{c}{\textbf{ECG200}} \\
\toprule
 & \multicolumn{1}{c|}{\textbf{Single}} & \multicolumn{1}{c|}{\textbf{Ensemble}} &   \multicolumn{1}{c|}{\textbf{Single}} & \multicolumn{1}{c}{\textbf{Ensemble}} \\
\hline
k=5 & .80 $\pm$ .07 & \textbf{.93 $\pm$ 0.02} & .57 $\pm$ .15 & \textbf{.82 $\pm$ 0.07} \\
\hline
k=7 & .85 $\pm$ .05 & \textbf{.94 $\pm$ 0.02} & .60 $\pm$ .15 & \textbf{.84 $\pm$ 0.07} \\
\hline
k=10 & .89 $\pm$ .04 & \textbf{.95 $\pm$ 0.01} & .63 $\pm$ .14 & \textbf{.85 $\pm$ 0.06} \\
\hline
k=15 & .91 $\pm$ .04 & \textbf{.97 $\pm$ 0.01} & .67 $\pm$ .13 & \textbf{.88 $\pm$ 0.05} \\
\bottomrule
\end{tabular}
\label{tab:Consistency}
\end{table}

\subsection{Myocardial Infarction Detection}
\begin{figure*}[t]
\centering
  \subfloat[\label{fig:ecgMrel}]{%
       \includegraphics[width=0.90\linewidth]{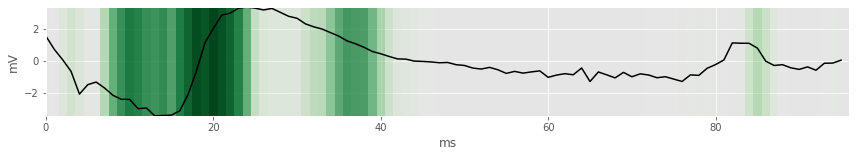}}
    \hfill
  \subfloat[\label{fig:ecgSrel}]{%
       \includegraphics[width=0.90\linewidth]{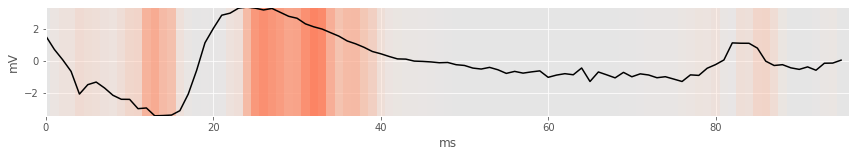}}
    \hfill
  \subfloat[\label{fig:ecgMSrel}]{%
       \includegraphics[width=0.90\linewidth]{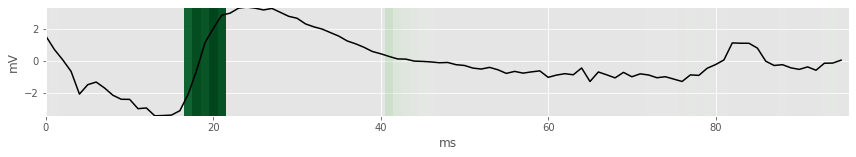}}
    \hfill
\caption{Example from the myocardial infarction class that was correctly classified by the deep ensemble. From top to bottom: mean relevance scores across all models in ensemble (a), standard deviation across all models in ensemble (b), and uncertainty filtered relevance scores obtained using the proposed method in Section \ref{sec:UFRS} (c). (a) shows that there are several regions of relevant time steps, but (b) indicates that there is a degree of uncertainty associated with several of those regions. (c) shows only the certainly relevant samples, where the uncertain time steps are filtered out using the proposed methodology.}
\label{fig:UcECG}
\end{figure*}

The proposed approach for measuring uncertainty in the relevance scores is validated on the ECG200 data described above. The performance of a single FCN and an ensemble of FCNs on the test data of the ECGO200 dataset, each trained for 150 epochs, is displayed in Table \ref{tab:metrics}. A Monte Carlo permutation test with 10000 permutations is conducted to test for significance. Results show that the ensemble is more capable of identifying positive samples. Figure \ref{fig:ecgMrel} and \ref{fig:ecgSrel} show the mean and standard deviation of the relevance scores across all models in the ensemble for a myocardial infarction case. This sample was correctly classified by the ensemble as belonging to the myocardial infarction class. Figure \ref{fig:ecgMrel} indicates that there are three regions of electrical activity that influenced the prediction of the model. First, a steep incline in the initial ST-period, second a slight decline after a peak, and lastly, a peak towards the end. However, Figure \ref{fig:ecgSrel} shows that there are several regions where the models in the ensemble disagrees on the relevance of different time steps. By using the proposed method of uncertainty filtered relevance scores,  Figure \ref{fig:ecgMSrel} is created. Here, the uncertain relevance scores are removed and only the certain scores remain. Now, Figure \ref{fig:ecgMSrel} shows that there is only one region of highly certain and relevant time steps, which is the initial incline in electrical activity in the ST-period. This rapid change is electrical activity is also associated with the myocardial infarction class \cite{olszewski2001generalized}, which suggests that the ensemble is able to capture clinically relevant features in the input data.

\begin{figure*}[t]
\centering
  \subfloat[\label{fig:ecgSingle}]{%
       \includegraphics[width=0.90\linewidth]{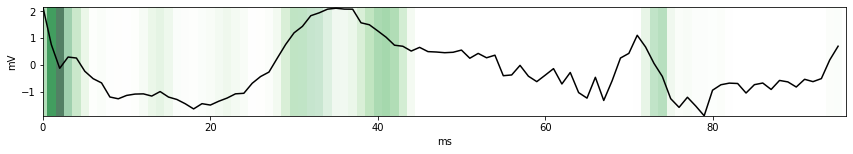}}
    \hfill
  \subfloat[\label{fig:ecgEnsemble}]{%
       \includegraphics[width=0.90\linewidth]{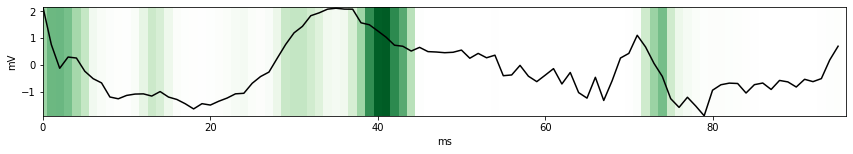}}
    \hfill
\caption{Example for the heart attack class where a single model fails to detect the heart attack while the ensemble correctly classifies the patient. The figure shows the relevance scores for the single model (a) and the ensemble model (b) for the prediction of the heart attack class. While there are similarities between the relevance scores, the single model puts more emphasis on some initial time steps while the ensemble focuses on time steps related to the ST-period.}
\label{fig:SingVsEns}
\end{figure*}

Lastly, we present an example where the ensemble classifies a sample correctly as a heart attack while the single model makes an error. Figure \ref{fig:SingVsEns} displays the relevance scores for both the single and ensemble model for this particular example. While the relevance scores have similarities, notice that the single model emphasis the importance of some initial time steps, while the ensemble model indicates that the decline after the ST-period is the most relevant part of the time series. The single model also indicate this part of the time series as important, but less important than the initial part of the time series. This shows how the ensemble is capable of filtering out time steps that might not be important and focus on clinically relevant features, which in this case leads to a correct classification.

\subsection{Surgical Site Infection Detection}
\begin{figure}
    \centering
    \includegraphics[width=0.975\linewidth]{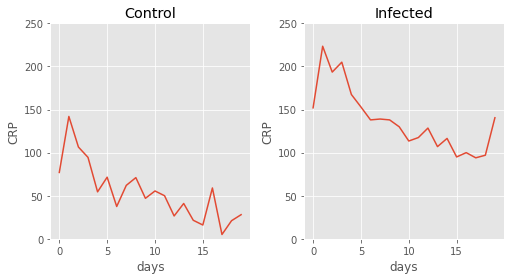}
    \caption{Median CRP at each time steps for each class of all samples in the training dataset. Figure shows how control patients usually have an increase in CRP after surgery but declines to a low value towards the end of the time series. The infected patients typically have a higher value of CRP and also tend to have an increase in CRP towards the end of the time series.}
    \label{fig:CversusI}
\end{figure}

\begin{figure*}[t]
\centering
  \subfloat[\label{fig:ssiMrel}]{%
       \includegraphics[width=0.95\linewidth]{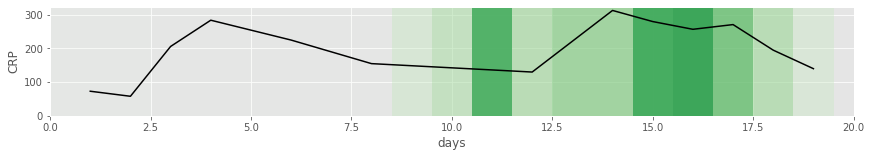}}
    \hfill
  \subfloat[\label{fig:ssiSrel}]{%
       \includegraphics[width=0.95\linewidth]{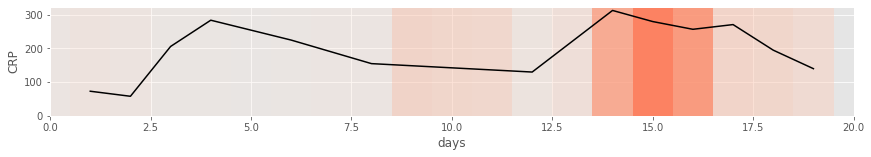}}
    \hfill
  \subfloat[\label{fig:ssiMSrel}]{%
       \includegraphics[width=0.95\linewidth]{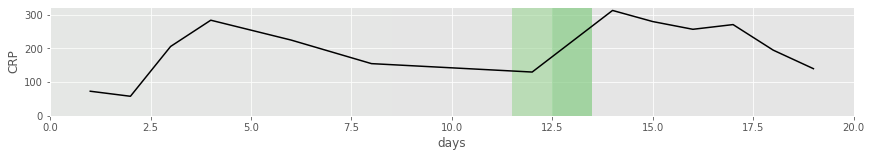}}
    \hfill
\caption{Example from the infection class that was correctly classified by the deep ensemble. From top to bottom: mean relevance scores across all models in ensemble (a), standard deviation across all models in ensemble (b), and uncertainty filtered relevance scores obtained using the proposed method in Section \ref{sec:UFRS} (c). (a) shows that there are several time steps highlighted as important for the prediction, but (b) shows that there is a degree of uncertainty associated with several of them. (c) shows that the only certainly relevant time steps are those associated with the CRP incline around day 12-13, a pattern known to correlated with the risk of contracting an infection.}
\label{fig:UcSSI}
\end{figure*}

The next task the proposed methodology is validated on is surgical site infection in measurements of CRP. The dataset consists of 883 patients that have undergone a gastrointestinal surgical procedure at the Department of Gastrointestinal Surgery at the University Hospital of North Norway in the years 2004 - 2012 \cite{Mikalsen2016LearningSB}. Of the 883 patients, 232 are infected while the rest are control patients. 80 \% of the data were used for training and 20 \% were used as an independent test set. This split is conducted at each independent training run to obtain a cross-validated evaluation of performance. The performance of a single FCN and an ensemble of FCNs on the test data of the SSI dataset, each trained for 150 epochs, is displayed in Table \ref{tab:metrics}. A Monte Carlo permutation test with 10000 permutations is conducted to test for significance. Results display that the ensemble has higher precision compared to single models.

Figure \ref{fig:UcSSI} displays an example of a patient that contracted an infection and was correctly classified by the deep ensemble. Figure \ref{fig:ssiMrel} and \ref{fig:ssiSrel} show the mean and standard deviation of the relevance scores across all models in the ensemble. Figure \ref{fig:ssiMrel} indicates that there are several time step deemed relevant by the ensemble, and particularly the rise in CRP is indicated as highly relevant for the prediction. However, Figure \ref{fig:ssiSrel} shows that a number of these these time steps have a high degree of uncertainty associated with them. Particularly, the relevance of the central plateau and final parts of the CRP measurement is something that the models in the ensemble disagree on. Figure \ref{fig:ssiMSrel} displays the standard deviation filtered mean relevance scores, which indicates that only the incline around day 13 is the only certainly relevant part for the prediction of infection for this patient.

The typical development of CRP for a patient that has undergone surgery but does not contract an infection is an initial postoperative increase followed by a steady decline. For patients that do get an infection, CRP typically sees an increase again some days postoperatively after the initial decline. The correlation between CRP and the risk of infection has been noted in previous studies \cite{crpRef, cristina}. Figure \ref{fig:CversusI} shows the median CRP at each time steps for each class of all samples in the training dataset. This illustrates how CRP is typically higher for infected patients, and they tend to have an increase in CRP after the initial decline after surgery. Figure \ref{fig:ssiMSrel} shows the incline at 12-13 days after surgery are indicated as the certainly relevant time steps for the prediction of the surgical site infection class. As described in this paragraph, such a pattern is closely connected with a possibility of developing an infection, which suggests that the deep ensemble uses clinically relevant features to make its prediction.

\section{Conclusion}
In this work a deep ensemble approach for explainable CNNs was proposed. The proposed method was evaluated on both synthetic and real world data. Results demonstrate that deep ensembles are capable of finding relevant features in clinical time series and that by modeling the uncertainty in relevance scores more understandable and trustworthy explanations can be provided. A novel thresholding approach was proposed and demonstrated. While only one thresholding was investigated in this work, we believe that different thresholding strategies could be applicable, which is an interesting line of research for future works. The contributions of this work can enable the construction of explainable decision support systems that are more trustworthy and more accurate than previous systems based on deep learning.

\section*{Acknowledgment}
The authors would like to thank Jonas Nordhaug Myhre, Luigi Tomasso Luppino, and Ahcene Boubekki for helpful discussions during both the conception and development of this work.

% Can use something like this to put references on a page
% by themselves when using endfloat and the captionsoff option.
\ifCLASSOPTIONcaptionsoff
  \newpage
\fi

% trigger a \newpage just before the given reference
% number - used to balance the columns on the last page
% adjust value as needed - may need to be readjusted if
% the document is modified later
%\IEEEtriggeratref{8}
% The "triggered" command can be changed if desired:
%\IEEEtriggercmd{\enlargethispage{-5in}}

\bibliographystyle{IEEEtran}
\bibliography{bibliography}

\end{document}